# The Role of Prompt Language and Translation-Theory-Driven Prompts in Large Language Models: A Case Study on Spanish–Chinese Journalistic Translation

**Haohong Lai**
Faculty of Translation and Interpreting
Autonomous University of Barcelona
Barcelona, Spain
haohong.lai@autonoma.cat

**Weijia Li**
Faculty of Translation and Interpreting
Autonomous University of Barcelona
Barcelona, Spain
weijia.li@autonoma.cat

## Abstract

This study examines how prompt language and translation theory-driven prompt design influence the quality of Spanish–Chinese journalistic translations generated by GPT-5.2. A parallel corpus of four editorials from EL PAÍS was translated under 48 experimental conditions (4 prompt types × 3 prompt languages × 4 articles). Translation quality was assessed using BLEU and BERTScore-F1 for automated evaluation, alongside human evaluation based on the Multidimensional Quality Metrics (MQM) framework. Automated metrics identified the baseline prompt (BASE) as the best-performing condition, whereas human evaluation ranked the brief-oriented prompt (BRIEF) highest (MQM: 8.66 vs. 7.84), a reversal likely attributable to the single-reference constraint inherent in automated measures. Sub-error type analysis revealed that translation-theory-driven prompts selectively reduced Awkward style errors, while Unidiomatic style errors persisted across conditions. Prompt language had a negligible impact under both evaluation paradigms. These results indicate that translation-theory-driven prompts can yield measurable quality gains under expert evaluation of journalistic translations, although their pedagogical implications for language learners remain suggestive and require validation through user-based studies.

## 1 Introduction

The academic consensus has long held that engagement with newspapers and periodicals constitutes a widespread and significant means of language acquisition (Lee and Morrison, 1998). More recently, generative artificial intelligence tools, including models such as ChatGPT, have gained prominence as resources for language learning and textual comprehension (Chatterji et al., 2025; Liu et al., 2025). These tools are promoted not only by AI developers but also increasingly examined and endorsed in scholarly research. Empirical studies indicate that large language models (LLMs) can deliver translation quality comparable to, and in some cases surpassing, that of conventional machine translation systems across diverse language pairs and text genres (Bhattacharjee et al., 2024; Briva-Iglesias et al., 2024; Eschbach-Dymanus et al., 2024; Hendy et al., 2023; Jiao et al., 2023; Moreno García and Mangiron, 2024; Sanz-Valdivieso and López-Arroyo, 2023). Against this backdrop, a growing number of language learners are turning to AI-powered tools to translate foreign-language publications and support their language learning processes (Zapata, 2025).

LLMs are increasingly recognised for their potential to reshape professional translation practices. This development marks a shift from conventional tool-centric workflows towards more adaptive human–computer interactions driven by natural-language instructions (Sánchez-Gijón and Palenzuela-Badiola, 2023). With continuing advances in prompt engineering, the crucial influence of prompt design on the quality of LLM-generated translation is now well established. Although carefully designed prompts have been shown to enhance output quality (Cao et al., 2025; Laki and Yang, 2024), the indiscriminate application of certain prompt types across different models can impair translation performance (Peng et al., 2023), resulting in inefficient use of computational resources and higher operational costs (Iglesias and Dogru, 2025). Accordingly, translators need to critically evaluate and strategically employ different prompting approaches if LLMs are to be integrated effectively into translation practice (Yamada, 2023).

Numerous studies have sought to improve translation performance through the development of a range of prompting strategies (Gao et al., 2024; Hendy et al., 2023; Laki and Yang, 2024; Mondshine et al., 2025; Peng et al., 2023; Vilar et al., 2023; Zhang et al., 2023; Zhu et al., 2024). Nonetheless, the bulk of this work has been conducted by researchers specialising in natural language processing (NLP), who typically have limited formal training in translation studies. As a result, its methodological framework is predominantly shaped by conventional zero-shot and few-shot learning approaches. This tendency may be attributed in part to the disciplinary gap between translation studies and computer science. By contrast, translation scholars who seek to engage in prompt engineering often face a steep learning curve in acquiring the necessary technical expertise.

Several recent studies have sought to integrate linguistic knowledge into prompt design (Gu, 2025; He, 2024; Kunilovskaya et al., 2024; Yamada, 2023). Nevertheless, such approaches, including chain-of-thought (CoT) and chain-of-dictionary prompting strategies (Lu et al., 2024), often entail considerable complexity and tend to overlook the increased cognitive load placed on non-expert users who employ LLMs for translation. Consequently, current research and practice in translation prompt engineering rarely address the specific needs of language learners using AI-powered translation tools, despite the clear importance of this user group.

By ‘translation-theory-driven prompts’ we refer to prompts that explicitly encode constructs drawn from translation studies. In this study, we operationalise three such constructs: (i) translator subjectivity (Venuti, 2017; Yu and Yao, 2026), realised through an expert-role assignment; (ii) register and text typology (Pym, 2023; Reiss, 2000), realised by providing the publication context of the source text; and (iii) skopos and translation brief (He, 2024; Nord, 2014), realised through an integrated brief specifying purpose, audience, and situational context. These three constructs map directly onto the ROLE, CTX, and BRIEF prompt types, respectively, with BASE serving as the theory-free control.

Grounded in the actual needs of language learners and informed by key concepts from translation theory, the present study formulates tailored translation prompts and addresses the following research questions:

- RQ1: Do translation-theory-driven prompts improve translation quality?
- RQ2: Does the language of the prompt affect translation quality?

## 2 Literature Review

### 2.1 Zero-shot and few-shot prompt design

A common strategy in research aimed at enhancing the quality of LLM-generated translations through prompt engineering involves incorporating a limited number of reference translations into the prompt. This approach utilises the model’s in-context learning capabilities (Brown et al., 2020). Within this paradigm, the quality of the examples is regarded as the most critical factor (Vilar et al., 2023). When provided with high-quality exemplars, few-shot prompts have consistently outperformed one-shot prompts (Zhang et al., 2023). A substantial body of research has demonstrated the effectiveness of few-shot prompting across a range of domains and language pairs, including specialised corpora in technology, medicine, news, and idioms (Castaldo and Monti, 2024; Gao et al., 2024; Laki and Yang, 2024; Peng et al., 2023). Moreover, the use of a rudimentary few-shot prompt structure comprising translation instructions alone has been shown to significantly improve output quality (Gao et al., 2024). Finally, setting the temperature parameter to zero has been shown to improve output stability, particularly for languages with complex grammatical structures, such as Chinese (Peng et al., 2023).

While few-shot prompting has been shown to yield competitive translation quality for high-resource language pairs, its limitations remain evident in low-resource settings (Hendy et al., 2023). Zhu et al. (2024) corroborated this pattern in a large-scale evaluation encompassing 606 language directions. For high-resource pairs, GPT-4 with eight-shot prompts demonstrated translation quality comparable to NLLB [1]. However, when applied to low-resource languages, its performance exhibited a substantial decline. In a similar vein, systematic experiments by Toukmaji (2024) showed that, for low-resource languages such as Kinyarwanda, Hausa, and Luganda, few-shot prompting did not consistently outperform pivot prompting (Jiao et al., 2023) or model fine-tuning across downstream tasks (Stap and Araabi, 2023). In some cases, few-shot prompting even underperformed relative to simple majority-class baselines.

A number of studies have investigated the potential of few-shot prompting beyond the direct integration of reference translations into prompt examples. For instance, Lee et al. (2023) employed few-shot prompts to direct LLMs to generate high-quality translation samples as synthetic data for training machine translation models. Their study demonstrates the effectiveness of this approach. Qian and Kong (2024) integrated few-shot prompting with CoT prompting to ascertain whether LLMs possess the capacity to identify particular linguistic concepts and their relevant contexts, and to generate translations that reflect these characteristics.

## 2.2 Other prompt design

In addition, several studies have investigated methods for enhancing the quality of LLM-generated translation without relying on few-shot prompting techniques. For example, Wu and Hu (2023) found that engaging GPT in multi-turn, dialogue-based translation requests improved performance for the Chinese–English language pair. Similarly, Yang et al. (2023) demonstrated that translation quality can be enhanced by enabling GPT to automatically retrieve relevant content from memory and incorporate human-provided reference translation feedback into the prompt.

## 2.3 Integrating disciplinary knowledge into prompt design

Several studies have incorporated interdisciplinary knowledge, particularly from the field of linguistics, into prompt design. For instance, Kunilovskaya et al. (2024) incorporated linguistic knowledge into translation prompts to guide GPT-4 in revising English and German translations so as to mitigate foreignisation. Experimental results indicate that the revised outputs displayed more natural target-language phrasing while preserving source-text meaning, thereby reducing surface-level translationese. Gu (2025) similarly proposed a three-step prompt-chain strategy that integrated linguistic features to help ChatGPT accurately translate Japanese relative clauses into Chinese. The findings of both studies demonstrate the effectiveness of incorporating linguistic expertise into LLM elicitation.

A limited number of studies have also incorporated translation theory into prompt design. Yamada's (2023) investigation sought to ascertain the impact of incorporating translation purpose and target-audience information into prompts on the quality of translations produced by ChatGPT. The findings suggest that explicitly specifying the purpose of the translation leads to outputs that are more culturally aligned with the target audience, thereby facilitating intentional translation and domestication. He's (2024) study most closely parallels the prompt-design methodology adopted in the present study. In her proposal, He (2024) draws on translation theory to propose two types of prompt. The first is a translation brief, which integrates contextual information and textual functionality. The second is a role-based prompt, which assigns ChatGPT specific identities, such as "translator" or "author". The findings indicate that basic prompts exhibit superior performance compared with translated-brief prompts. Moreover, assigning ChatGPT the role of translator leads to optimal outcomes.

## 2.4 Prompt language

Existing evidence suggests that prompt language affects LLM performance. In the context of machine translation, English prompt templates demonstrate average performance. However, when translating into Chinese, Chinese templates can yield improvements (Zhang et al., 2023). In cross-lingual multitask settings, selectively translating prompt components into English generally outperforms both the "all-English" and "all-source-language" approaches (Mondshine et al., 2025).

## 2.5 Research Gaps

[1] NLLB (No Language Left Behind) is a supervised multi-lingual neural machine translation model developed by Meta AI, capable of translating between 202 languages. Its strong performance—particularly on low-resource languages—makes it a widely adopted benchmark in multilingual MT research.

However, a review of previous studies reveals that the Spanish–Chinese language pair remains largely unexplored. The existing literature typically compares the effectiveness of zero-shot and few-shot prompting strategies. By contrast, prompt designs that incorporate linguistic or theoretical knowledge are often overly complex and offer limited practical guidance for non-expert users, who, in the context of this study, are foreign language learners. In addition, existing research has not adequately addressed the effect of prompt language. Contemporary research frequently employs English as the default prompt language, which is closely related to the dominant role of English in intercultural communication. However, most users prefer to interact with AI products in their native language. Therefore, this study conceptualises prompt language as a crucial variable in the evaluation process. The case study examined here focuses on the translation of Spanish newspaper editorials into Chinese. The native languages of potential end users may include Chinese, which is likely to be the majority language, Spanish, or English, while other language backgrounds are also possible. The case therefore offers limited but useful insights into how prompt language may operate across different user language backgrounds.

## 3 Experimental Setup

### 3.1 Prompt Design

The prompt design in this study is primarily informed by key concepts derived from translation theory. Translation, as a complex cross-cultural communicative act (Venuti, 2000), is governed by multiple constraints and objectives.

Translation studies offer many candidate concepts that could inform prompt design. We have selected three concepts: translator subjectivity (Venuti, 2017), register/text typology (Pym, 2023; Reiss, 2000), and skopos (Nord, 2014; Vermeer, 2000) based on the following criteria: (a) direct operationalisability in a single-turn prompt, (b) correspondence to established prompt-engineering variables identified in prior work (He, 2024; Yamada, 2023), and (c) parsimony: construct maps onto one distinct prompt element, avoiding the cognitive load of multi-component strategies such as chain-of-thought or chain-of-dictionary prompting (Lu et al., 2024).

First, translators are not invisible agents; their subjectivity should be acknowledged throughout the translation process (Venuti, 2017). Translators bring personal experiences, knowledge, and perspectives that inevitably influence their choices. Consequently, different translators may render the same source text in different ways, guided by their professional expertise, cultural background, and ethical stance. For example, a translator with a strong literary background may prioritise a text's aesthetic qualities, whereas one with technical expertise may emphasise accuracy and clarity in specialised terminology. Cultural background also plays a critical role: a translator from a Western context may interpret certain concepts differently from one in an Eastern context, resulting in variation in the translated text. Ethical stance further shapes translation decisions; some practitioners may adhere closely to the source text, while others may adapt more freely to enhance accessibility for the target audience. As He (2024) observes, assigning a specific translator identity to an LLM can affect both its stylistic tendencies and output quality. Accordingly, this study's prompt templates assign the LLM the role of a specialist in journalistic translation.

From the perspectives of text typology and register analysis, the extratextual context of a source text critically shapes its meaning and informs the choice of translation strategies (Pym, 2023; Reiss, 2000). Register analysis examines how language varies according to field (topic), tenor (relationship among participants), and mode (communication medium). For example, a legal document requires a highly formal register and precise rendering of terms, whereas an informal conversation permits more colloquial and flexible phrasing. Extrinsic factors, including historical period, social circumstances, and cultural milieu, provide crucial clues to a text's intent. A work produced during a momentous historical event may contain references that resonate only within that particular context. Equipping LLMs with rich contextual information about the source text allows them to discern its communicative purpose and intended meaning more accurately, thereby minimising literal, decontextualised translations and culturally inappropriate interpretations.

The prompt design also reflects two environmental and computational-efficiency considerations. First, prompts are kept concise to minimise unnecessary token redundancy, since even minor inefficiencies in token utilisation accumulate into substantial computational overhead as model use scales (Ligozat et al., 2022;

Rillig et al., 2023; Yin et al., 2024). Second, prompts employ neutral or moderately polite phrasing combined with clear task constraints, in line with evidence that such formulations are more effective for LLM instruction-following than either strongly polite or strongly impolite alternatives (Dobariya and Kumar, 2025; Yin et al., 2024).

Building on these principles, four prompt types were developed in each of three languages (Chinese, Spanish, and English): a task-only baseline (BASE), which served as the control condition; an expert-role prompt (ROLE), which specifies the translator identity; a context-primed prompt (CTX), which provides the publication context of the source text; and a brief-oriented prompt (BRIEF), which combines role and context into a compact translational brief (Chesterman, 2016; Nord, 2014). Prompt templates are provided in Appendix A.

### 3.2 Text Type and Data

In this study, a small parallel corpus was compiled from four open-access editorials published in the *Opinión* section of the renowned Spanish newspaper EL PAÍS. Excel was used to organise, segment, and align the data. It is important to note that this parallel corpus was not constructed on a sentence-by-sentence basis (Rothwell et al., 2023). Instead, it was segmented according to the original paragraph structure of the EL PAÍS editorials. This approach is based on two considerations. First, contemporary LLMs possess strong contextual understanding capabilities (Brown et al., 2020) and are therefore likely to produce more coherent and natural translations at the segment level. Second, in professional translation practice, translators typically process source texts sentence by sentence with the aid of translation support systems, a sentence-level mode of operation determined by the working mechanisms of the software involved. However, when foreign language learners use machine translation to access information, they may not follow the same processing workflow (Nurminen, 2025). To take full advantage of AI models' capacity for contextual modelling, or for reasons of efficiency, they may prefer to understand and process texts as whole paragraphs or even larger semantic units.

Each editorial ranges from 460 to 648 words (an average of 557 words), and is structured as four paragraphs. A readability analysis using the Flesch-Szigriszt index with the INFLESZ scale (Barrio-Cantalejo et al., 2008) places all four texts in the 'Muy difícil (Very difficult)' to 'Difícil (Difficult)' band (mean IFSZ = 46.5, SD = 9.9; range 33–57), consistent with reader profiles at CEFR C1–C2, that is, the target user profile of the study: advanced language learners accessing foreign-language press. Full per-article metadata is provided in Appendix B.

### 3.3 Environmental Configuration

All translations were generated using GPT-5.2, accessed via the OpenAI API. The temperature[2] parameter was set to 0 for all conditions to ensure deterministic output, following Peng et al. (2023), who report that this setting improves output stability for morphologically complex target languages, including Chinese. No system-level commands[3] were used; all experimental variables were encoded in the user prompt. Each of the 48 article-level runs (4 prompt types × 3 prompt languages × 4 articles) was submitted as a single API call with the full source-text passage included.

### 3.4 Evaluation Framework

Translation quality was assessed through two complementary paradigms: automatic reference-based evaluation and human evaluation using the Multidimensional Quality Metrics (MQM) framework. The rationale for combining the two lies in the documented divergence between surface-form metrics and expert human judgements, particularly for target languages requiring a specific register (Freitag et al., 2021; Kocmi et al., 2021).

Automatic evaluation was performed using the sacrebleu library (Post, 2018) and the bert-score library (Zhang et al., 2020). BLEU was computed at both the corpus level and the sentence-average level, using character-level tokenisation (tokenize = 'zh') in accordance with standard WMT practice for Chinese. BERTScore-F1 was computed using *bert-base-chinese* with baseline rescaling; precision and recall are not reported separately, as F1 is the accepted convention in MT evaluation. For each of the 48 article-level runs, metrics were first computed across all paragraph segments of the article. The resulting scores were then averaged across articles within each prompt type × prompt language combination and subsequently summarised by prompt type and by prompt language.

[2] Temperature is a sampling parameter that controls output randomness: lower values make outputs more focused and deterministic, whereas higher values increase variability (Dubois et al., 2025).

[3] System- or developer-level commands refer to higher-priority API instructions supplied outside the user message.

Human evaluation followed the MQM framework (Lommel et al., 2014), using a two-stage workflow. In the first stage, two bilingual annotators with professional experience in Spanish–Chinese translation independently evaluated all 48 conditions, comparing the model output against the source text while using a professional reference translation as a register benchmark. Conditions were evaluated in random order to minimise systematic bias. In the second stage, the annotators held adjudication meetings to resolve disagreements in error identification, error-type classification, and severity assignment, producing a single consensus MQM record per condition. The consensus scores reported below are derived from this adjudicated dataset, not from averages of independent scorings. Per-condition MQM scores were computed as follows: $MQM = \max(0, 10 + \Sigma(w_s \times n_s))$, where w_s denotes the severity weight (Minor = −0.1, Medium = −0.3, Major = −0.5) and n_s is the number of errors at each severity level.

## 4 Results

This section presents the results of both evaluation paradigms applied to all 48 translation outputs generated across four EL PAÍS editorials, four prompt types (BASE, ROLE, CTX, BRIEF), and three prompt languages (ZH, ES, EN). The automatic evaluation results are presented in Section 4.1. Section 4.2 presents the results of the human evaluation using the MQM framework, covering overall quality scores, error subtype analysis, and segment-level quality distributions.

### 4.1 Automatic evaluation

Two metrics are reported: (i) BLEU (corpus-level and sentence-averaged), computed using *sacrebleu* with tokenize = ‘zh’, following standard WMT practice for Chinese (Post, 2018); and (ii) BERTScore-F1, computed using *bert-base-chinese* with baseline rescaling (Zhang et al., 2020). For each article-level run, metric scores were first computed across all segments of the article. These scores were then averaged across the four articles within each prompt type × prompt language combination.

| Prompt type | BLEU-corpus | BLEU-sent.avg | BERTScore-F1 |
|---|---|---|---|
| BASE | **48.80** | **48.02** | **0.7241** |
| CTX | 48.37 | **48.02** | 0.7201 |
| ROLE | 48.28 | 47.34 | 0.7142 |
| BRIEF | 46.59 | 46.17 | 0.7052 |

**Table 1**. Automatic evaluation results by prompt type (mean across 4 articles × 3 prompt languages; N = 12 per condition). Boldface indicates the best value in each column. BLEU was computed using sacrebleu with tokenize = ‘zh’; BERTScore-F1 was computed using *bert-base-chinese* with baseline rescaling.

**Effect of prompt type**. Table 1 presents automatic evaluation scores by prompt type, averaged across four articles and three prompt languages. BASE achieves the highest BERTScore-F1 (0.7241) and BLEU-corpus (48.80), while BRIEF obtains the lowest values on both metrics (BERTScore-F1: 0.7052; BLEU-corpus: 46.59). CTX and ROLE occupy intermediate positions. The maximum between-condition differences are small (ΔBERTScore-F1 = 0.019; ΔBLEU-corpus = 2.21), and the ranking BASE > CTX > ROLE > BRIEF is consistent across all four articles.

| Prompt language | BLEU-corpus | BLEU-sent.avg | BERTScore-F1 |
|---|---|---|---|
| ES | **48.44** | **47.78** | **0.7198** |
| EN | 47.98 | 47.27 | 0.7150 |
| ZH | 47.61 | 47.11 | 0.7129 |

**Table 2**. Automatic evaluation results by prompt language (mean across 4 articles × 4 prompt types; N = 16 per language). Boldface indicates the best value in each column.

**Effect of prompt language**. Table 2 reports automatic evaluation scores by prompt language, averaged across four articles and four prompt types. Spanish-language prompts achieve the highest BLEU-corpus (48.44) and BERTScore-F1 (0.7198), followed by English (47.98; 0.7150) and Chinese (47.61; 0.7129). Differences are negligible across all metrics (ΔBERTScore-F1 = 0.007; ΔBLEU-corpus = 0.83).

| Prompt type | MQM score | Total errors | Accuracy | Linguistic conv. | Style | |
|---|---|---|---|---|---|---|
| | | | (total) | (total) | (total) | (major) |
| BASE | 7.84 | 7.9 | **0.0** | 1.6 | 6.3 | 2.0 |
| ROLE | 8.35[a] | 7.4 | 0.3 | 1.4 | 5.8 | 0.9 |
| CTX | 8.30[a] | 7.0 | 0.3 | 1.5 | 5.2 | 1.3 |
| BRIEF | **8.66**[abc] | **6.5** | 0.3 | **1.3** | **4.9** | **0.5** |

**Table 3**. MQM human evaluation results by prompt type (mean per condition; N = 12 per prompt type). Higher MQM scores and lower error counts indicate better performance. Boldface indicates the best value in each column. Superscripts indicate statistically significant pairwise differences in MQM score after Holm-Bonferroni correction: [a] p_adj < .01 vs. BASE; [b] p_adj < .05 vs. ROLE; [c] p_adj < .05 vs. CTX. ROLE and CTX do not differ significantly. See Appendix C for full Wilcoxon signed-rank test results.

| Prompt language | MQM score | Total errors | Accuracy | Linguistic conv. | Style | |
|---|---|---|---|---|---|---|
| | | | (total) | (total) | (total) | (major) |
| ZH | 8.20 | 7.3 | 0.2 | **1.3** | 5.8 | 1.4 |
| EN | **8.33** | **7.2** | 0.2 | 1.5 | 5.5 | **1.0** |
| ES | **8.33** | **7.2** | 0.2 | 1.6 | **5.4** | 1.2 |

**Table 4**. MQM human evaluation results by prompt language (mean per condition; N = 16 per language). No pairwise contrast between languages reaches statistical significance after Holm-Bonferroni correction (all p_adj > .17; see Appendix C).

| Category | Subtype | BASE | ROLE | CTX | BRIEF | Total |
|---|---|---|---|---|---|---|
| Style | Unidiomatic style | **38** | 40 | 41 | 40 | 159 |
| | Awkward style | 38 | 29 | 22 | **19** | 108 |
| | Subtotal | 76 | 69 | 63 | **59** | 267 |
| Linguistic conv. | Punctuation | 16 | 14 | 15 | **13** | 58 |
| | Grammar | 3 | 3 | 3 | 3 | 12 |
| | Subtotal | 19 | 17 | 18 | **16** | 70 |
| Accuracy | Mistranslation | **0** | 3 | 3 | 3 | 9 |
| Total | | 95 | 89 | 84 | **78** | 346 |

**Table 5**. MQM error counts by subtype and prompt type (raw totals across all 48 conditions).

In summary, automatic metrics consistently rank BASE highest and BRIEF lowest across both BLEU and BERTScore, with prompt language exerting a negligible effect.

## 4.2 Human evaluation

**Effect of prompt type on MQM score**. Table 3 presents MQM scores and average errors per condition by prompt type. BRIEF achieves the highest mean MQM score (8.66), followed by ROLE (8.35), CTX (8.30), and BASE (7.84). This ranking is the exact inverse of the automatic evaluation ranking reported in §4.1. The gap between BRIEF and BASE is 0.82 MQM points, a difference that is statistically robust and corresponds to a fourfold reduction in Major Style errors.

The most pronounced difference across prompt types lies in Major Style errors: BASE conditions average 2.0 Major Style errors per condition, compared with 0.5 for BRIEF, representing a fourfold reduction. ROLE (0.9) and CTX (1.3) occupy intermediate positions. Total Style errors also decrease monotonically from BASE (6.3 per condition) to BRIEF (4.9). BRIEF records the fewest total errors (6.5 per condition), suggesting that source-text context may be most effective when combined with role, audience, and purpose information in an integrated translation brief. Linguistic convention errors remain broadly stable across prompt types (1.3-1.6 per condition), and Accuracy errors are absent from BASE but appear at a low rate (0.3 per condition) across the three theory-driven prompt types.

**Effect of prompt language on MQM score**. Table 4 reports MQM results by prompt language. English- and Spanish-language prompts achieve the same mean MQM score (8.33), with Chinese-language prompts trailing marginally (8.20). All differences are small: the maximum MQM range across languages is 0.13, and per-condition error counts differ by no more than 0.1 across ES and EN (both 7.2) versus ZH (7.3). The only dimension showing a somewhat clearer language effect is Major Style errors: Chinese prompts produce 1.4 Major Style errors per condition, compared with 1.0 for Spanish and 1.2 for English.

| Segment | BASE | ROLE | CTX | BRIEF | Avg. errors (all types) |
|---|---|---|---|---|---|
| Segment 1 (opening) | 9.44 | 9.50 | 9.55 | 9.64 | 1.88 |
| Segment 2 | 9.39 | 9.58 | 9.53 | 9.57 | 1.85 |
| Segment 3 | 9.69 | 9.74 | 9.72 | 9.78 | 1.44 |
| Segment 4 (closing) | 9.49 | 9.71 | 9.65 | 9.76 | 1.44 |

**Table 6**. Segment-level MQM scores by prompt type (mean across 4 articles × 3 prompt languages; N = 12 per cell).

Thus, under human evaluation, prompt type produces the inverse ranking of automatic metrics, while prompt language differences remain small in absolute terms (max 0.13 MQM points).

**Error subtype analysis**. Table 5 presents the full error subtype breakdown across all 48 conditions. Style errors constitute 77.2% of all annotated errors (267/346), followed by Linguistic conventions (20.2%; n = 70) and Accuracy (2.6%; n = 9). Within Style, Unidiomatic style is the most frequent subtype (n = 159; 59.6% of Style errors), followed by Awkward style (n = 108; 40.4%). Within Linguistic conventions, Punctuation predominates (n = 58; 82.85%). Accuracy errors consist exclusively of Mistranslation instances and are entirely absent from BASE conditions, appearing only in ROLE, CTX, and BRIEF.

The most informative cross-prompt pattern concerns Awkward style errors: counts decline markedly with increasing theoretical grounding in the prompts (BASE: 38 -> ROLE: 29 -> CTX: 22 -> BRIEF: 19), while Unidiomatic style error counts remain essentially constant across prompt types (38-41 per type). This dissociation indicates that theory-driven prompts selectively reduce locally ill-formed stylistic constructions while leaving the model's tendency towards non-native idiomatic choices unresolved. The subtype data reveal a dissociation: theory-driven prompts selectively attenuate locally ill-formed (Awkward) stylistic constructions while leaving idiomatic (Unidiomatic) naturalness essentially unchanged.

**Segment-level quality distribution**. Table 6 presents segment-level MQM scores by prompt type, averaged across all 48 conditions (12 per cell). Each of the four articles consists of four paragraphs (Segments 1-4 in document order). Two findings emerge. First, error density is somewhat higher in the opening segments: Segments 1 and 2 average 1.88 and 1.85 errors per condition, respectively, compared with 1.44 each for Segments 3 and 4. This pattern is consistent across all prompt types and suggests that the opening paragraphs of EL PAÍS editorials, which typically introduce the central argument and establish the stylistic register, present a greater translation challenge than the concluding paragraphs. Second, the quality advantage of translation-theory-driven prompts is present at every segment position without exception, confirming that prompt-type effects are not localised to any particular position in the document.

## 5 Discussion

The findings presented in Section 4 yield two central observations. First, automatic metrics and human evaluation produce inverse prompt-type rankings, with important implications for how LLM translation quality should be assessed in journalistic translation contexts. Second, prompt language exerts only a marginal effect under both evaluation paradigms. These observations are discussed in turn with reference to the two research questions.

### 5.1 RQ1: Do translation-theory-driven prompts improve translation quality?

The MQM results provide a qualified positive answer to RQ1. The BRIEF prompt, grounded in Nord's (2014) functionalist skopos framework and Chesterman's (2016) concept of the translation brief, achieves the highest MQM score (8.66) and the fewest Major Style errors (0.5 per condition). The ROLE prompt ranks second (8.35), followed by CTX (8.30). The improvement from BASE to BRIEF represents a gain of 0.82 points on a 10-point scale, while the fourfold reduction in Major Style errors, from 2.0 to 0.5 per condition, suggests that brief-oriented prompts are associated with a substantial decrease in severe stylistic failures when translating Spanish editorials into Chinese.

The error subtype data further refine this finding. The progressive decline in Awkward style errors across prompt types (BASE: 38 > ROLE: 29 > CTX: 22 > BRIEF: 19) indicates that translation-theory-driven prompts primarily reduce locally ill-formed stylistic constructions, that is, those identifiable at clause or sentence level. By contrast, Unidiomatic style errors remain stable at approximately 40 per prompt type, suggesting that GPT-5.2's underlying tendency to produce non-idiomatic Chinese editorial phrasing is not addressed by prompt design alone. This distinction can be interpreted in terms of a difference between broader register framing, which BRIEF appears to improve in this dataset, and fine-grained lexical-idiomatic naturalness, which no prompt type in the present study fully resolves.

The automatic metrics reverse this ranking entirely. BASE produces the highest BERTScore-F1 (0.7241) and BLEU-corpus (48.80), while BRIEF scores lowest (0.7052; 46.59). This inversion reflects the single-reference evaluation problem (Freitag et al., 2021; Kocmi et al., 2021): when a brief-oriented prompt directs the model towards a register or stylistic framing different from that of the reference translation, the resulting output incurs BLEU and BERTScore penalties because it deviates from the reference surface form, while at the same time being communicatively more appropriate for the intended readership. This effect is particularly consequential for the language-learning scenario that motivates this study: foreign language learners accessing EL PAÍS editorials are better served by translations that conform to Chinese editorial register conventions than by translations that maximise surface similarity to a single professionally produced reference. These findings suggest that automatic metrics should not be used as the sole basis for evaluating LLM prompting strategies in journalistic translation tasks.

The segment-level analysis adds a further dimension to these findings. The consistent advantage of translation-theory-driven prompts across all four segment positions confirms that prompt-type effects are not an artefact of any particular section of the document. The higher error density in the opening segments (1.85-1.88 errors per condition) relative to the closing segments (1.44 each) may suggest that GPT-5.2 benefits from the accumulation of prior target-language context during generation, thereby producing more coherent and register-consistent output as it processes longer portions of the source text.

### 5.2 RQ2: Does the language of the prompt affect translation quality?

The results provide little evidence that prompt language meaningfully affects translation quality. English- and Spanish-language prompts achieve the same mean MQM score (8.33), with Chinese-language prompts trailing only marginally (8.20; $\Delta$ = 0.13). Differences in the automatic metrics across prompt languages are similarly negligible ($\Delta$BERTScore-F1 = 0.007). No language-related advantage is detectable at the segment level at any position in the document.

The absence of a target-language (Chinese) advantage is practically relevant for the intended user group of this study. Chinese-speaking foreign language learners might naturally prefer to compose prompts in their native language; the present data indicate that this preference incurs no measurable quality cost compared with prompting in Spanish or English. GPT-5.2's cross-lingual instruction-following appears sufficiently robust for prompt language not to constitute a performance bottleneck for this language pair and task type, consistent with findings on multilingual prompting for high-resource language pairs (Ahuja et al., 2023; Shi et al., 2022).

The marginally elevated Major Style error count for Chinese-language prompts (1.4 vs. 1.0-1.2 per condition) runs counter to the target-language hypothesis and is consistent with a slight interference effect between Chinese meta-instructions and Chinese target-text generation. However, the effect is small, and its interpretation requires caution given the absence of statistically significant language-related differences. The practical recommendation for language learners is clear: prompt structure, specifically whether the prompt conveys a translational brief, matters substantially more than prompt language. This recommendation, however, concerns translation quality rather than efficiency, since prompt language may still affect token consumption and token-based cost through language- and tokenizer-specific differences (Komatsuzaki, 2026; Lundin et al., 2026).

### 5.3 Limitations

Several limitations of this study should be acknowledged. First, the empirical scope is necessarily narrow. The dataset consists of only four editorials from the Opinión section of EL

PAÍS, all translated in a single direction, from Spanish into Simplified Chinese. While this design allows for a tightly controlled comparison across prompt types and prompt languages, it limits the generalisability of the findings to other genres, language pairs, and translation purposes. In particular, the results may not transfer directly to literary, technical, or legal texts, where discourse structure, terminological density, and register expectations differ substantially from those of newspaper editorials. Likewise, the study examines only one model, GPT-5.2, under a fixed configuration (temperature = 0), so the observed effects may be model-specific rather than generally characteristic of LLM-based translation.

Second, the human evaluation followed a two-stage independent-plus-adjudication workflow that produced a single consensus record per condition. A consequence of this workflow is that the independent pre-adjudication scorings were not retained in a form that permits post-hoc computation of inter-annotator agreement statistics. Future applications of this protocol should pre-register agreement measures, such as, Cohen's κ for categorical decisions (error category, subtype) and weighted κ or intraclass correlation for ordinal severity ratings, and preserve both the independent and consensus scorings for reliability analysis.

Finally, although this research is motivated by the needs of foreign language learners, it does not include direct user-based validation. The conclusions are based on translation quality assessment rather than on learner comprehension, usability, prompting behaviour, or task efficiency in authentic learning settings. The pedagogical implications presented here should therefore be understood as provisional. Future research should extend the design to larger and more diverse corpora, additional LLMs and source texts, multi-reference and reliability-tested evaluation protocols, and user-centred experiments involving actual language learners.

## 6 Conclusion

This study examines the impact of prompt language and translation theory-driven prompt design on LLM-generated Spanish–Chinese journalistic translation, addressing the specific needs of foreign language learners who use AI tools to access press content. Two findings are of particular interest.

First, translation-theory-driven prompts improve translation quality as assessed by human evaluators. The BRIEF prompt, grounded in skopos theory and the translation brief concept, produced the highest MQM scores and the fewest Major Style errors across all four source texts, with consistent gains at every segment position in the document. This improvement was not captured, and was in fact reversed, by automatic metrics, confirming that reference-based metrics are insufficient for evaluating LLM translations in contexts where multiple register-appropriate renderings are acceptable.

Second, for this language pair and task type, prompt language had no meaningful effect on translation quality in this dataset. Prompts written in Chinese, Spanish, and English produced near-identical results under both evaluation paradigms, indicating that users may prompt in whichever language is most natural to them without any loss of quality. However, as noted in the discussion of RQ2, this finding concerns translation quality rather than efficiency: changes in prompt language may still affect token consumption and, by extension, token-based costs.

These findings connect with a growing literature on AI-mediated language learning (Lee et al., 2026). Recent studies document that learners increasingly use LLM-based tools for text comprehension and vocabulary expansion in L2 contexts (Chatterji et al., 2025; Pym and Hao, 2025; Zapata, 2025). Our results contribute a prompt-design perspective to this literature: the quality of LLM-produced reading support depends not only on tool choice but on how learners formulate instructions. This has implications for machine translation literacy in language education (Bowker and Buitrago Ciro, 2019; Loock and Léchauguette, 2021): prompt formulation skills (Hwang et al., 2023), such as specifying purpose, audience, and register, may warrant explicit treatment in advanced L2 curricula as a transferable machine translation literacy component. Rigorous pedagogical validation remains, however, beyond the scope of the present quality-assessment design.

Taken together, these results suggest a tentative recommendation which should be empirically validated in user-based studies, that prompt structure, rather than prompt language, is the variable most worthy of focused attention. Specifically, providing the model with a brief that specifies target audience, publication context, and stylistic register yields measurable quality gains that variation in prompt language does not. Future work should extend this investigation to other LLMs, additional source languages, other publication types, and multi-annotator MQM protocols that enable reliability assessment.

## Carbon impact statement

This study involved inference-only use of GPT-5.2 via the OpenAI API and did not include any model training or fine-tuning. The experiment comprised 48 article-level translation runs (4 prompt types × 3 prompt languages × 4 articles), followed by automatic evaluation with BLEU and BERTScore and human evaluation using MQM. Given the small experimental scale and the absence of training, the overall computational and carbon impact of the study is expected to be limited when compared with training-intensive or large-scale benchmarking studies.

## References

Kabir Ahuja, Harshita Diddee, Rishav Hada, Millicent Ochieng, Krithika Ramesh, Prachi Jain, Akshay Nambi, Tanuja Ganu, Sameer Segal, Mohamed Ahmed, Kalika Bali, and Sunayana Sitaram. 2023. MEGA: Multilingual Evaluation of Generative AI. In Houda Bouamor, Juan Pino, and Kalika Bali, editors, *Proceedings of the 2023 Conference on Empirical Methods in Natural Language Processing*, pages 4232–4267, Singapore. Association for Computational Linguistics.

Inés María Barrio-Cantalejo, P. Simón-Lorda, M. Melguizo, I. Escalona, M. I. Marijuán, and P. Hernando. 2008. Validación de la Escala INFLESZ para evaluar la legibilidad de los textos dirigidos a pacientes. *Anales del Sistema Sanitario de Navarra*, 31(2):135–152.

Soham Bhattacharjee, Baban Gain, and Asif Ekbal. 2024. Domain Dynamics: Evaluating Large Language Models in English-Hindi Translation. In Sobha Lalitha Devi and Karunesh Arora, editors, *Proceedings of the 21st International Conference on Natural Language Processing (ICON)*, pages 169–177, AU-KBC Research Centre, Chennai, India. NLP Association of India (NLPAI).

Lynne Bowker and Jairo Buitrago Ciro. 2019. Towards a Framework for Machine Translation Literacy. In *Machine Translation and Global Research: Towards Improved Machine Translation Literacy in the Scholarly Community*, pages 87–95. Emerald Publishing Limited, Bingley.

Vicent Briva-Iglesias, Gokhan Dogru, and João Lucas Cavalheiro Camargo. 2024. Large language models “ad referendum”: How good are they at machine translation in the legal domain? *MonTI. Monographs in Translation and Interpreting*(16):75–107.

Tom Brown, Benjamin Mann, Nick Ryder, Melanie Subbiah, Jared D Kaplan, Prafulla Dhariwal, Arvind Neelakantan, Pranav Shyam, Girish Sastry, Amanda Askell, Sandhini Agarwal, Ariel Herbert-Voss, Gretchen Krueger, Tom Henighan, Rewon Child, Aditya Ramesh, Daniel Ziegler, Jeffrey Wu, Clemens Winter, et al. 2020. Language Models are Few-Shot Learners. In *Advances in Neural Information Processing Systems*, volume 33, pages 1877–1901. Curran Associates, Inc.

Peng Cao, Masood Khoshsaligheh, and Fatemeh Jomhouri. 2025. Deepseek, ChatGPT, and Gemini versus human subtitling: a case study in socio-cultural adaptation in multimedia communication. *Perspectives: Studies in Translation Theory and Practice*:1–19.

Antonio Castaldo and Johanna Monti. 2024. Prompting Large Language Models for Idiomatic Translation. In Vanroy, Bram, Lefer, Marie-Aude, Macken, Lieve, and Ruffo, Paola, editors, *Proceedings of the 1st Workshop on Creative-text Translation and Technology*, pages 37–44, Sheffield, UK. European Association for Machine Translation (EAMT).

Aaron Chatterji, Thomas Cunningham, David J. Deming, Zoe Hitzig, Christopher Ong, Carl Yan Shan, and Kevin Wadman. 2025.

How People Use ChatGPT. DOI: 10.3386/w34255.

Andrew Chesterman. 2016. *Memes of Translation: The Spread of Ideas in Translation Theory*.Benjamins Translation Library. John Benjamins, Amsterdam / Philadelphia, Revised edition.

Om Dobariya and Akhil Kumar. 2025. Mind Your Tone: Investigating How Prompt Politeness Affects LLM Accuracy. arXiv:2510.04950 [cs].

Matthieu Dubois, François Yvon, and Pablo Piantanida. 2025. How Sampling Affects the Detectability of Machine-written texts: A Comprehensive Study. In Christos Christodoulopoulos, Tanmoy Chakraborty, Carolyn Rose, and Violet Peng, editors, *Findings of the Association for Computational Linguistics: EMNLP 2025*, pages 11369–11387, Suzhou, China. Association for Computational Linguistics.

Johannes Eschbach-Dymanus, Frank Essenberger, Bianka Buschbeck, and Miriam Exel. 2024. Exploring the Effectiveness of LLM Domain Adaptation for Business IT Machine Translation. In *Proceedings of the 25th Annual Conference of the European Association for Machine Translation (Volume 1)*, pages 610–622, Sheffield, United Kingdom. European Association for Machine Translation (EAMT).

Markus Freitag, George Foster, David Grangier, Viresh Ratnakar, Qijun Tan, and Wolfgang Macherey. 2021. Experts, Errors, and Context: A Large-Scale Study of Human Evaluation for Machine Translation. *Transactions of the Association for Computational Linguistics*, 9:1460–1474.

Yuan Gao, Ruili Wang, and Feng Hou. 2024. How to Design Translation Prompts for ChatGPT: An Empirical Study. In *Proceedings of the 6th ACM International Conference on Multimedia in Asia Workshops*, pages 1–7, Auckland, New Zealand. Association for Computing Machinery.

Wenshi Gu. 2025. Linguistically informed ChatGPT prompts to enhance Japanese-Chinese machine translation: A case study on attributive clauses. *PLOS ONE*, 20(1):e0313264.

Sui He. 2024. Prompting ChatGPT for Translation: A Comparative Analysis of Translation Brief and Persona Prompts. In Carolina Scarton, Charlotte Prescott, Chris Bayliss, Chris Oakley, Joanna Wright, Stuart Wrigley, Xingyi Song, Edward Gow-Smith, Rachel Bawden, Víctor M Sánchez-Cartagena, Patrick Cadwell, Ekaterina Lapshinova-Koltunski, Vera Cabarrão, Konstantinos Chatzitheodorou, Mary Nurminen, Diptesh Kanojia, and Helena Moniz, editors, *Proceedings of the 25th Annual Conference of the European Association for Machine Translation*, volume 1, pages 316–326, Sheffield, UK. European Association for Machine Translation (EAMT).

Amr Hendy, Mohamed Abdelrehim, Amr Sharaf, Vikas Raunak, Mohamed Gabr, Hitokazu Matsushita, Young Jin Kim, Mohamed Afify, and Hany Hassan Awadalla. 2023. How Good Are GPT Models at Machine Translation? A Comprehensive Evaluation. arXiv:2302.09210 [cs].

Yohan Hwang, Jang Ho Lee, and Dongkwang Shin. 2023. What is prompt literacy? An exploratory study of language learners' development of new literacy skill using generative AI. arXiv:2311.05373 [cs].

Vicent Briva Iglesias and Gokhan Dogru. 2025. AI agents may be worth the hype but not the resources (yet): An initial exploration of machine translation quality and costs in three language pairs in the legal and news domains. arXiv:2505.01560 [cs].

Wenxiang Jiao, Wenxuan Wang, Jen-tse Huang, Xing Wang, Shuming Shi, and Zhaopeng Tu. 2023. Is ChatGPT A Good Translator? Yes With GPT-4 As The Engine. arXiv:2301.08745 [cs].

Tom Kocmi, Christian Federmann, Roman Grundkiewicz, Marcin Junczys-Dowmunt, Hitokazu Matsushita, and Arul Menezes.

2021. To Ship or Not to Ship: An Extensive Evaluation of Automatic Metrics for Machine Translation. In Loic Barrault, Ondrej Bojar, Fethi Bougares, Rajen Chatterjee, Marta R. Costa-jussa, Christian Federmann, Mark Fishel, Alexander Fraser, Markus Freitag, Yvette Graham, Roman Grundkiewicz, Paco Guzman, Barry Haddow, Matthias Huck, Antonio Jimeno Yepes, Philipp Koehn, Tom Kocmi, Andre Martins, Makoto Morishita, et al., editors, *Proceedings of the Sixth Conference on Machine Translation*, pages 478–494, Online. Association for Computational Linguistics.

Aran Komatsuzaki. 2026. The non-English tax is real. Sutton's Bitter Lesson, translated across languages and normalized to OpenAI English token count. X post, April 28, 2026. https://x.com/arankomatsuzaki/status/2049125048792006965. Accessed 30 April 2026.

Maria Kunilovskaya, Koel Dutta Chowdhury, Heike Przybyl, Cristina España-Bonet, and Josef Genabith. 2024. Mitigating Translationese with GPT-4: Strategies and Performance. In Carolina Scarton, Charlotte Prescott, Chris Bayliss, Chris Oakley, Joanna Wright, Stuart Wrigley, Xingyi Song, Edward Gow-Smith, Rachel Bawden, Víctor M Sánchez-Cartagena, Patrick Cadwell, Ekaterina Lapshinova-Koltunski, Vera Cabarrão, Konstantinos Chatzitheodorou, Mary Nurminen, Diptesh Kanojia, and Helena Moniz, editors, *Proceedings of the 25th Annual Conference of the European Association for Machine Translation (Volume 1)*, pages 411–430, Sheffield, United Kingdom. European Association for Machine Translation (EAMT).

László János Laki and Zijian Győző Yang. 2024. Improving Machine Translation Capabilities by Fine-Tuning Large Language Models and Prompt Engineering with Domain-Specific Data. In pages 1–5, Debrecen, Hungary. Institute of Electrical and Electronics Engineers (IEEE).

Seongyong Lee, Hohsung Choe, Di Zou, and Jaeho Jeon. 2026. Generative AI (GenAI) in the language classroom: A systematic review. *Interactive Learning Environments*, 34(1):335–359.

Seungjun Lee, Hyeonseok Moon, Chanjun Park, and Heuiseok Lim. 2023. Improving Formality-Sensitive Machine Translation Using Data-Centric Approaches and Prompt Engineering. In Elizabeth Salesky, Marcello Federico, and Marine Carpuat, editors, *Proceedings of the 20th International Conference on Spoken Language Translation (IWSLT 2023)*, pages 420–432, Toronto, Canada. Association for Computational Linguistics.

Winnie Lee and Bruce Morrison. 1998. A Role for Newspaper Articles in Developing Autonomous Language Learning Skills. *RELC Journal*, 29(2):90–120.

Anne-Laure Ligozat, Julien Lefevre, Aurélie Bugeau, and Jacques Combaz. 2022. Unraveling the Hidden Environmental Impacts of AI Solutions for Environment Life Cycle Assessment of AI Solutions. *Sustainability*, 14(9).

Chaoqun Liu, Wenxuan Zhang, Yiran Zhao, Anh Tuan Luu, and Lidong Bing. 2025. Is Translation All You Need? A Study on Solving Multilingual Tasks with Large Language Models. In Luis Chiruzzo, Alan Ritter, and Lu Wang, editors, *Proceedings of the 2025 Conference of the Nations of the Americas Chapter of the Association for Computational Linguistics: Human Language Technologies (Volume 1: Long Papers)*, pages 9594–9614, Albuquerque, New Mexico. Association for Computational Linguistics.

Arle Lommel, Hans Uszkoreit, and Aljoscha Burchardt. 2014. Multidimensional Quality Metrics (MQM): A Framework for Declaring and Describing Translation Quality Metrics. *Revista Tradumàtica. Tecnologies de la Traducció*(12):455–463.

Rudy Loock and Sophie Léchauguette. 2021. Machine translation literacy and undergraduate students in applied languages: Report on an exploratory study. *Revista*

*Tradumàtica. Tecnologies de la Traducció*(19):204–225.

Hongyuan Lu, Haoran Yang, Haoyang Huang, Dongdong Zhang, Wai Lam, and Furu Wei. 2024. Chain-of-Dictionary Prompting Elicits Translation in Large Language Models. In Yaser Al-Onaizan, Mohit Bansal, and Yun-Nung Chen, editors, *Proceedings of the 2024 Conference on Empirical Methods in Natural Language Processing*, pages 958–976, Miami, Florida, USA. Association for Computational Linguistics.

Jessica M. Lundin, Ada Zhang, Nihal Karim, Hamza Louzan, Guohao Wei, David Ifeoluwa Adelani, and Cody Carroll. 2026. The Token Tax: Systematic Bias in Multilingual Tokenization. In Everlyn Asiko Chimoto, Constantine Lignos, Shamsuddeen Muhammad, Idris Abdulmumin, Clemencia Siro, and David Ifeoluwa Adelani, editors, *Proceedings of the 7th Workshop on African Natural Language Processing (AfricaNLP 2026)*, pages 103–112, Rabat, Morocco. Association for Computational Linguistics.

Itai Mondshine, Tzuf Paz-Argaman, and Reut Tsarfaty. 2025. Beyond English: The Impact of Prompt Translation Strategies across Languages and Tasks in Multilingual LLMs. In Atul Kr. Ojha, Chao-hong Liu, Ekaterina Vylomova, Flammie Pirinen, Jonathan Washington, Nathaniel Oco, and Xiaobing Zhao, editors, *Proceedings of the Eighth Workshop on Technologies for Machine Translation of Low-Resource Languages (LoResMT 2025)*, pages 81–104, Albuquerque, New Mexico, U.S.A. Association for Computational Linguistics.

Luis Damián Moreno García and Carme Mangiron. 2024. Exploring the potential of GPT-4 as an interactive transcreation assistant in game localisation: A case study on the translation of Pokémon names. *Perspectives: Studies in Translation Theory and Practice*:1–18.

Christiane Nord. 2014. *Translating as a Purposeful Activity: Functionalist Approaches Explained*. Routledge, 2nd ed.

Keqin Peng, Liang Ding, Qihuang Zhong, Li Shen, Xuebo Liu, Min Zhang, Yuanxin Ouyang, and Dacheng Tao. 2023. Towards Making the Most of ChatGPT for Machine Translation. In *Findings of the Association for Computational Linguistics: EMNLP 2023*, pages 5622–5633, Singapore. Association for Computational Linguistics.

Matt Post. 2018. A Call for Clarity in Reporting BLEU Scores. In Ondřej Bojar, Rajen Chatterjee, Christian Federmann, Mark Fishel, Yvette Graham, Barry Haddow, Matthias Huck, Antonio Jimeno Yepes, Philipp Koehn, Christof Monz, Matteo Negri, Aurélie Névéol, Mariana Neves, Matt Post, Lucia Specia, Marco Turchi, and Karin Verspoor, editors, *Proceedings of the Third Conference on Machine Translation: Research Papers*, volume 1, pages 186–191, Brussels, Belgium. Association for Computational Linguistics.

Anthony Pym. 2023. *Exploring Translation Theories*. Routledge, 3rd ed.

Anthony Pym and Yu Hao. 2025. *How to Augment Language Skills: Generative AI and Machine Translation in Language Learning and Translator Training*. Routledge.

Ming Qian and Chuiqing Kong. 2024. Enabling Human-Centered Machine Translation Using Concept-Based Large Language Model Prompting and Translation Memory. In Helmut Degen and Stavroula Ntoa, editors, *Artificial Intelligence in HCI*, pages 118–134, Washington, D.C. Springer Nature Switzerland.

Katharina Reiss. 2000. Type, Kind and Individuality of Text: Decision making in translation. In Lawrence Venuti and Mona Baker, editors, *The Translation Studies Reader*. Routledge, London.

Matthias C. Rillig, Marlene Ågerstrand, Mohan Bi, Kenneth A. Gould, and Uli Sauerland. 2023. Risks and Benefits of Large

Language Models for the Environment. *Environmental Science & Technology*, 57(9):3464–3466.

Andrew Rothwell, Joss Moorkens, María Fernández-Parra, Joanna Drugan, and Frank Austermuehl. 2023. *Translation Tools and Technologies*. Routledge, 1st ed.

Pilar Sánchez-Gijón and Leire Palenzuela-Badiola. 2023. Analysis And Evaluation of ChatGPT-Induced HCI Shifts in the Digitalised Translation Process. In *Proceedings of the International Conference Human-informed Translation and Interpreting Technology (HiT-IT 2023)*, pages 227–267, Naples, Italy.

Lucía Sanz-Valdivieso and Belén López-Arroyo. 2023. Google Translate vs. ChatGPT: Can non-language professionals trust them for specialized translation? In *Proceedings of the International Conference on Human-informed Translation and Interpreting Technology 2023*, pages 97–107. INCOMA Ltd., Shoumen, Bulgaria.

Freda Shi, Mirac Suzgun, Markus Freitag, Xuezhi Wang, Suraj Srivats, Soroush Vosoughi, Hyung Won Chung, Yi Tay, Sebastian Ruder, Denny Zhou, Dipanjan Das, and Jason Wei. 2022. Language Models are Multilingual Chain-of-Thought Reasoners. arXiv:2210.03057 [cs].

David Stap and Ali Araabi. 2023. ChatGPT is not a good indigenous translator. In *Proceedings of the Workshop on Natural Language Processing for Indigenous Languages of the Americas (AmericasNLP)*, pages 163–167, Toronto, Canada. Association for Computational Linguistics.

Francisco Szigriszt Pazos. 1992. *Sistemas predictivos de legilibilidad del mensaje escrito: fórmula de perspicuidad*. Tesis Doctorales, Universidad Complutense de Madrid, Madrid.

Christopher Toukmaji. 2024. *Few-Shot Cross-Lingual Transfer for Prompting Large Language Models in Low-Resource Languages*. Undergraduate thesis, University of California, Santa Cruz. arXiv:2403.06018 [cs].

Lawrence Venuti. 2000. Translation, Communication, Utopia. In Lawrence Venuti and Mona Baker, editors, *The Translation Studies Reader*, pages 468–488. Routledge, London.

Lawrence Venuti. 2017. *The Translator's Invisibility: A History of Translation*. Routledge, London.

Hans J. Vermeer. 2000. Skopos and Commission in Translational Action. In Lawrence Venuti and Mona Baker, editors, *The Translation Studies Reader*, pages 221–233. Routledge, London.

David Vilar, Markus Freitag, Colin Cherry, Jiaming Luo, Viresh Ratnakar, and George Foster. 2023. Prompting PaLM for Translation: Assessing Strategies and Performance. In *Proceedings of the 61st Annual Meeting of the Association for Computational Linguistics (Volume 1: Long Papers)*, volume 1, pages 15406–15427, Toronto, Canada. Association for Computational Linguistics.

Yangjian Wu and Gang Hu. 2023. Exploring Prompt Engineering with GPT Language Models for Document-Level Machine Translation: Insights and Findings. In Philipp Koehn, Barry Haddow, Tom Kocmi, and Christof Monz, editors, *Proceedings of the Eighth Conference on Machine Translation (WMT)*, pages 166–169, Singapore. Association for Computational Linguistics.

Masaru Yamada. 2023. Optimizing Machine Translation through Prompt Engineering: An Investigation into ChatGPT's Customizability. In Masaru Yamada and Felix do Carmo, editors, *Proceedings of Machine Translation Summit XIX, Vol. 2: Users Track*, volume 2, pages 195–204, Macau SAR, China. Asia-Pacific Association for Machine Translation.

Xinyi Yang, Runzhe Zhan, Derek F. Wong, Junchao Wu, and Lidia S. Chao. 2023. Human-in-the-loop Machine Translation with Large Language Model. In Masaru Yamada and Felix do Carmo, editors, *Proceedings of Machine Translation Summit XIX, Vol. 2: Users Track*, volume 2, pages 88–98, Macau SAR, China. Asia-Pacific Association for Machine Translation.

Ziqi Yin, Hao Wang, Kaito Horio, Daisuke Kawahara, and Satoshi Sekine. 2024. Should We Respect LLMs? A Cross-Lingual Study on the Influence of Prompt Politeness on LLM Performance. In James Hale, Kushal Chawla, and Muskan Garg, editors, *Proceedings of the Second Workshop on Social Influence in Conversations (SICon 2024)*, pages 9–35, Miami, Florida, USA. Association for Computational Linguistics.

Jingsong Yu and Yazhi Yao. 2026. *Intelligent Language Services: Theory and Practice with Large Language Models*. Springer Nature Singapore, Singapore.

Gabriela C. Zapata. 2025. Introduction: Generative AI Technologies in Language Education: What We Know So Far. In Gabriela C. Zapata, editor, *Generative AI Technologies, Multiliteracies, and Language Education*, Multiliteracies and Second Language Education, pages 1–13. Routledge, London.

Biao Zhang, Barry Haddow, and Alexandra Birch. 2023. Prompting Large Language Model for Machine Translation: A Case Study. In *Proceedings of the 40th International Conference on Machine Learning*, pages 41092–41110, Honolulu, Hawaii, United States. PMLR.

Tianyi Zhang, Varsha Kishore, Felix Wu, Kilian Q. Weinberger, and Yoav Artzi. 2020. BERTScore: Evaluating Text Generation with BERT. arXiv:1904.09675 [cs].

Wenhao Zhu, Hongyi Liu, Qingxiu Dong, Jingjing Xu, Shujian Huang, Lingpeng Kong, Jiajun Chen, and Lei Li. 2024. Multilingual Machine Translation with Large Language Models: Empirical Results and Analysis. In Kevin Duh, Helena Gomez, and Steven Bethard, editors, *Findings of the Association for Computational Linguistics: NAACL 2024*, pages 2765–2781, Mexico City, Mexico. Association for Computational Linguistics.

**Appendix A. Prompt Templates**

| Prompt type | Language | Prompt content |
|---|---|---|
| BASE (Baseline) | ZH | 请将以下西班牙语文本翻译成简体中文：[待翻译文本] |
| | ES | Por favor, traduce el siguiente texto del español al chino simplificado: [Texto a traducir] |
| | EN | Please translate the following Spanish text into Simplified Chinese: [Text to translate] |
| ROLE (Expert Role) | ZH | 你是一位资深的西班牙语新闻翻译专家，专注于社论和深度评论文章的翻译。请利用你的专业知识，将以下文本翻译成准确、流畅且风格得体的简体中文：[待翻译文本] |
| | ES | Eres un experto traductor de noticias en español, especializado en la traducción de editoriales y artículos de opinión. Por favor, utiliza tu experiencia para traducir el siguiente texto a un chino simplificado preciso, fluido y estilísticamente apropiado: [Texto a traducir] |
| | EN | You are an expert Spanish news translator, specializing in editorials and opinion pieces. Please use your expertise to translate the following text into accurate, fluent, and stylistically appropriate Simplified Chinese: [Text to translate] |
| CTX (Context Primed) | ZH | {具体语境}<br>请结合这一语境，将以下文本翻译成简体中文：[待翻译文本] |
| | ES | {Contexto Específico}<br>Por favor, traduce el siguiente texto al chino simplificado teniendo en cuenta este contexto: [Texto a traducir] |
| | EN | {Specific Context}<br>Please translate the following text into Simplified Chinese, taking into account this context: [Text to translate] |
| BRIEF<br>(Brief Oriented)<br>(Integrates Role & Context) | ZH | 你是一位资深的西班牙语新闻翻译专家，专注于社论和深度评论文章的翻译。请根据以下翻译简报完成任务：<br>1. 文章背景：{具体语境}<br>2. 目标受众：关注国际时事的中国读者。<br>3. 翻译目的：准确传达作者的观点、立场及论证细节，供读者参考。<br>请将以下文本翻译成简体中文：[待翻译文本] |
| | ES | Eres un experto traductor de noticias en español, especializado en la traducción de editoriales y artículos de opinión. Por favor, completa la tarea siguiendo este encargo de traducción:<br>1. Contexto: {Contexto Específico}<br>2. Audiencia meta: Lectores chinos interesados en la actualidad internacional.<br>3. Propósito: Transmitir con precisión la opinión, la postura y los argumentos del autor para referencia del lector.<br>Traduce el siguiente texto al chino simplificado: [Texto a traducir] |
| | EN | You are an expert Spanish news translator, specializing in editorials and opinion pieces. Please complete the task according to the following translation brief:<br>1. Context: {Specific Context}<br>2. Target Audience: Chinese readers interested in international current affairs.<br>3. Purpose: To accurately convey the author's opinion, stance, and detailed arguments for the reader's reference.<br>Please translate the following text into Simplified Chinese: [Text to translate] |

**Table A1**. Prompt templates by type.

| Article | Language | {Specific Context} Content |
|---|---|---|
| Golpe a los abusos de Meta[4] | ZH | 这篇文章是西班牙《国家报》针对科技巨头 Meta 滥用市场支配地位发表的社论。背景涉及欧盟对数字平台的监管收紧、反垄断罚款及对用户隐私的保护。 |
| | ES | Este texto es un editorial de EL PAÍS sobre los abusos de posición dominante del gigante tecnológico Meta. El contexto implica el endurecimiento de la regulación de la UE sobre plataformas digitales, multas antimonopolio y la protección de la privacidad. |
| | EN | This text is an editorial from EL PAÍS regarding the abuse of market dominance by the tech giant Meta. The context involves tighter EU regulation of digital |

[4] Article link: https://elpais.com/opinion/2025-11-21/golpe-a-los-abusos-de-meta.html.

| | | |
|---|---|---|
| | | platforms, antitrust fines, and data privacy protection. |
| Dudas sobre los derechos digitales[5] | ZH | 这篇文章是《国家报》关于数字时代公民权利保护面临不确定性的社论。背景涉及新技术（如 AI 或监控系统）的快速发展与现行法律框架之间的冲突，以及对个人自由的担忧。 |
| | ES | Este texto es un editorial de EL PAÍS sobre la incertidumbre en la protección de los derechos de los ciudadanos en la era digital. El contexto abarca el conflicto entre el rápido avance de nuevas tecnologías (como la IA o la vigilancia) y los marcos legales actuales. |
| | EN | This text is an editorial from EL PAÍS concerning the uncertainty surrounding the protection of civil rights in the digital age. The context involves the conflict between rapid technological advances (such as AI or surveillance) and current legal frameworks. |
| Reformar y reforzar las pensiones[6] | ZH | 这篇文章是《国家报》讨论西班牙养老金体系改革与强化的社论。背景涉及在人口老龄化趋势下，如何通过政策调整确保公共财政的可持续性及社会福利的公平性。 |
| | ES | Este texto es un editorial de EL PAÍS que analiza la reforma y el fortalecimiento del sistema de pensiones en España. El contexto gira en torno a cómo garantizar la sostenibilidad fiscal y la equidad social frente al envejecimiento demográfico. |
| | EN | This text is an editorial from EL PAÍS discussing the reform and reinforcement of the pension system in Spain. The context centers on ensuring fiscal sustainability and social equity amidst an aging population. |
| Tormenta en el coste de la vida[7] | ZH | 这篇文章是《国家报》针对生活成本危机（通胀风暴）发表的社论。背景涉及物价普遍上涨对普通家庭经济造成的严峻压力，以及对购买力下降的社会焦虑。 |
| | ES | Este texto es un editorial de EL PAÍS sobre la crisis del coste de la vida (tormenta inflacionaria). El contexto refleja la severa presión que la subida generalizada de precios ejerce sobre la economía familiar y la ansiedad social por la pérdida de poder adquisitivo. |
| | EN | This text is an editorial from EL PAÍS addressing the cost-of-living crisis (inflationary storm). The context reflects the severe pressure that widespread price hikes place on household economics and the social anxiety regarding the loss of purchasing power. |

**Table A2**. {Specific Context} descriptions by article.

[5] Article link: https://elpais.com/opinion/2025-11-24/dudas-sobre-los-derechos-digitales.html.
[6] Article link: https://elpais.com/opinion/2025-11-29/reformar-y-reforzar-las-pensiones.html.
[7] Article link: https://elpais.com/opinion/2025-11-17/tormenta-en-el-coste-de-la-vida.html.

## Appendix B. Source-Text Metadata

| Article | Words | Sen-tences | Sylla-bles | Words/sent. | IFSZ | INFLESZ grade |
|---|---|---|---|---|---|---|
| Golpe a los abusos de Meta | 460 | 16 | 958 | 28.8 | 48 | Very difficult |
| Dudas sobre los derechos digitales | 544 | 18 | 1,254 | 30.2 | 33 | Very difficult |
| Reformar y reforzar las pensiones | 577 | 19 | 1,194 | 30.4 | 48 | Very difficult |
| Tormenta en el coste de la vida | 648 | 22 | 1,253 | 29.5 | 57 | Difficult |
| Mean | 557 | 18.75 | 1,165 | 29.7 | 46.5 | |

**Table B1**. Per-article source-text metadata. Readability indices were computed using Calculadora legibilidad Flesch[8], which implements the Flesch-Szigriszt formula (Szigriszt Pazos, 1992) with the INFLESZ grade scale (Barrio-Cantalejo et al., 2008). The INFLESZ scale is defined as: Very difficult (<40), Somewhat difficult (40–55), Normal (55–65), Quite easy (65–80), Very easy (>80).

## Appendix C. Pairwise Statistical Tests

Pairwise comparisons were performed using the Wilcoxon signed-rank test, a non-parametric alternative to the paired t-test appropriate for small samples and non-normal distributions. Effect size is reported as the matched-pairs rank-biserial correlation (r), where r = 1 indicates that the second condition outperformed the first on every paired observation. P-values were corrected for multiple comparisons across all nine contrasts (six prompt-type, three prompt-language) using the Holm-Bonferroni procedure.

For prompt-type comparisons, observations were paired by (article × prompt language), yielding N = 12 paired observations per contrast. For prompt-language comparisons, observations were paired by (article × prompt type), yielding N = 16 paired observations per contrast.

| Contrast | Δ mean | W | Raw p | Adj. p (Holm) | r |
|---|---|---|---|---|---|
| BRIEF − BASE | +0.81 | 0.0 | .0005 | **.0045** | 1.00 |
| ROLE − BASE | +0.48 | 0.0 | .0005 | **.0045** | 1.00 |
| CTX − BASE | +0.46 | 1.0 | .0010 | **.0070** | 0.97 |
| BRIEF − CTX | +0.35 | 2.5 | .0020 | **.0120** | 0.97 |
| BRIEF − ROLE | +0.33 | 3.5 | .0039 | **.0195** | 0.94 |
| ROLE − CTX | +0.03 | 34.0 | .7266 | .8434 | −0.11 |

**Table C1**. Prompt-type contrasts (N = 12). Bold = significant at α = .05 after Holm correction.

| Contrast | Δ mean | W | Raw p | Adj. p (Holm) | r |
|---|---|---|---|---|---|
| ZH − ES | +0.13 | 29.0 | .0434 | .174 | 0.57 |
| ZH − EN | +0.11 | 38.0 | .1203 | .361 | 0.46 |
| ES − EN | −0.03 | 52.5 | .4217 | .843 | −0.24 |

**Table C2**. Prompt-language contrasts (N = 16). No prompt-language contrast reached significance after Holm correction.

[8] For more information: https://creadorkit.com/seo-texto/calculadora-legibilidad-flesch/.